\documentclass[review]{elsarticle}
\usepackage[left=3.5cm, right=3.5cm, top=3cm]{geometry}

\usepackage{titlesec}
\titleformat{\section}
{\normalfont\Large\bfseries}%
{\thesection}{1em}{}
\titlespacing*{\section}
{0pt}{1.5\baselineskip}{1.5\baselineskip}

\titleformat{\subsection}
{\normalfont\large\bfseries}%
{\thesubsection}{1em}{}
\titlespacing*{\subsection}
{0pt}{1.5\baselineskip}{1.5\baselineskip}

\titleformat{\subsubsection}
{\normalfont\bfseries}%
{\thesubsubsection}{1em}{}
\titlespacing*{\subsubsection}
{0pt}{1.5\baselineskip}{1.\baselineskip}

\usepackage{multicol}
\usepackage[utf8]{inputenc}
\usepackage{siunitx}
\usepackage{comment}
\usepackage[toc,page,title]{appendix}

\usepackage{dsfont}
\usepackage{soul}
\usepackage{enumerate}
\usepackage[shortlabels]{enumitem}

\usepackage{amsmath,amsfonts,amsthm,amssymb}
\usepackage{empheq}
\usepackage{stmaryrd}
\usepackage[ruled,vlined]{algorithm2e}
\usepackage{mathalfa, eufrak, yfonts, mathrsfs}

\newtheorem{lemma}{Lemma}[subsection]

\newtheorem{theorem}{Theorem}[subsection]

\newtheorem{corollary}{Corollary}[subsection]

\theoremstyle{definition}
\newtheorem{definition}{Definition}[subsection]

\theoremstyle{definition}
\newtheorem*{definition*}{Definition}

\newtheorem{property}{Property}[subsection]

\newtheorem{example}{Example}[subsection]

\renewenvironment{proof}{{\bfseries Proof.}}{\qed}

\usepackage{amssymb,amsmath,amsfonts,amsthm} 
\usepackage{stmaryrd}
\usepackage[ruled,vlined]{algorithm2e}

\theoremstyle{definition}
\newtheorem*{remark}{Remark}





\newtheorem*{example*}{\small{Example}}
\usepackage{graphicx}
\graphicspath{ {./img/} }
\usepackage{xcolor}
\usepackage{comment}

\newcommand*\rulecirc[1]{%
	\begin{tikzpicture}
	\node[draw,circle,inner sep=0.5pt] {#1};
	\end{tikzpicture}
}

\usepackage[makeroom]{cancel}
\usepackage{pgfplots}
\usepackage{tikz}
\usetikzlibrary{decorations.pathreplacing}
\usetikzlibrary{positioning}
\usetikzlibrary{arrows}
\usetikzlibrary{arrows.meta}
\usetikzlibrary{shapes.geometric}
\usetikzlibrary{shapes.misc}
\tikzset{cross/.style={cross out, draw=red, minimum size=40*(#1-\pgflinewidth), inner sep=0pt, outer sep=0pt},
	cross/.default={1pt}}

\makeatletter
\newcommand{\osymbol}[1]{\mathbin{\mathpalette\make@circled#1}}
\newcommand{\make@circled}[2]{
	\ooalign{$\m@th#1\smallbigcirc{#1}$\cr\hidewidth$\m@th#1#2$\hidewidth\cr}
}
\newcommand{\smallbigcirc}[1]{
	\vcenter{\hbox{\scalebox{1}{$\m@th#1\bigcirc$}}}
}
\makeatother

\usepackage{pdfpages}
\usepackage{graphicx}
\usepackage{lineno,hyperref}
\modulolinenumbers[5]

\usepackage[export]{adjustbox}
\usepackage{subcaption}

\usepackage{parskip}
\usepackage[utf8]{inputenc}

\begin{document}

\begin{frontmatter}
	
	\title{
		Efficient exact computation of the conjunctive and disjunctive decompositions of D-S Theory for information fusion: Translation and extension\tnoteref{mytitlenote}}
	\tnotetext[mytitlenote]{This work was carried out and co-funded in the framework of the Labex MS2T and the Hauts-de-France region of France. It was supported by the French Government, through the program ``Investments for the future'' managed by the National Agency for Research (Reference ANR-11-IDEX-0004-02).}
	\author{Maxime Chaveroche$^*$, Franck Davoine, V\'eronique Cherfaoui}
	\ead{name.surname@hds.utc.fr}
	\address{Sorbonne University Alliance, Universit\'e de technologie de Compi\`egne, CNRS, Heudiasyc,\\
		CS 60319 - 60203 Compi\`egne Cedex, France}
	\cortext[mycorrespondingauthor]{Corresponding author}
	\begin{abstract}
		Dempster-Shafer Theory (DST) generalizes Bayesian probability theory, offering useful additional information, but suffers from a high computational burden. A lot of work has been done to reduce the complexity of computations used in information fusion with Dempster's rule.
		Yet, few research had been conducted to reduce the complexity of computations for the conjunctive and disjunctive decompositions of evidence, which are at the core of other important methods of information fusion. 
		In this paper, we propose a method designed to exploit the actual \textit{evidence} (information) contained in these decompositions in order to compute them. It is based on a new notion that we call \textit{focal point}, derived from the notion of focal set. With it, we are able to reduce these computations up to a linear complexity in the number of focal sets in some cases. In a broader perspective, our formulas have the potential to be tractable when the size of the frame of discernment exceeds a few dozen possible states, contrary to the existing litterature. 
		This article extends (and translates) our work published at the french conference GRETSI in 2019 \cite{me_gretsi}.
	\end{abstract}
\end{frontmatter}

\section{Introduction}

Dempster-Shafer Theory (DST) \cite{shafer76} is an elegant formalism that generalizes Bayesian probability theory. It is more expressive than the latter because it allows an observer to represent its beliefs in the state of a variable of interest not only by assigning credit directly to a specific state (strong evidence) but also by assigning credit to any set of possible states (weaker evidence). More precisely, for $\Omega$ the set of all possible outcomes, belief may be assigned to any set of $2^\Omega$ in DST.
This belief assignment is called a \textit{mass function} and provides meta-information quantifying the level of uncertainty of the observer itself, which is crucial for decision making.

Nevertheless, this information comes with a cost: considering $2^{|\Omega|}$ potential values may lead to temporally and spatially heavy algorithms. They can become difficult to use for more than a dozen possible states (e.g. 20 states in $\Omega$ generates about a million subsets), though one may need to consider more possible states (e.g. in classification or identification tasks).  
This complexity is even more limiting for real time applications. To tackle this issue, a lot of work has been done to reduce the complexity of computations used in information fusion with Dempster's rule \cite{dempster68}. 
We distinguish two main approaches that we refer to as \textit{powerset based} and \textit{evidence based}.

The \textit{powerset based} approach encompasses all algorithm based on the structure of the lattice $2^\Omega$. They have a complexity dependent on $|\Omega|$.
In the general case, the family of optimal algorithms of this kind is based on the \textit{Fast M\"obius Transform} (FMT) \cite{FMT}. Their complexity is $O(|\Omega|. 2^{|\Omega|})$ 
in time and $O(2^{|\Omega|})$ in space.

The \textit{evidence based} approach encompasses all algorithm that seeks to reduce computations to sets that carry information, namely \textit{focal sets}, which are in general far less numerous than $2^{|\Omega|}$. This approach is often more efficient than the powerset based one since it only depends on the information contained in sources with a quadratic complexity at most. Doing so, it enables one to exploit the full potential of DST by letting one choose any set of outcomes $\Omega$, no matter its size.

Furthermore, even if it is possible that this approach leads to situations in which the FMT is more efficient \cite{wilson2000}, these situations become less and less likely to occur as $|\Omega|$ grows.
In addition, the evidence based approach directly benefits from the use of information approximation methods, some of which being very efficient \cite{sarabi-jamab16}.
Thus, this approach seems superior to the FMT in most cases, above all when $|\Omega|$ is large, where a method with exponential complexity like the FMT is simply intractable.

The conjunctive decomposition, also known as \textit{canonical decomposition} \cite{conj_dec}, is an important belief representation in DST. In particular, it allowed for the definition of the \textit{Cautious fusion rule} \cite{disj_dec_cautious_bold} and its generalizations \cite{kallel09}. This rule consistently combines belief assignments that may not be independent, which is often the case in real applications such as information sharing in a vehicular network \cite{zoghby14}. Its dual, the disjunctive decomposition, has been proposed in \cite{disj_dec_cautious_bold} where it has been used to define the \textit{Bold fusion rule}. This latter rule is useful when sources may not be independent and are not entirely reliable. More generally, the conjunctive and disjunctive decompositions are used to define infinite families of t-norm and uninorm based fusion rules \cite{pichon2008}. They are also exploited in conflict analysis \cite{roquel2014}, clustering \cite{schubert2008clustering} and belief reinforcement/weakening \cite{
	mercier16}.

Yet, few options are available regarding the computation of the conjunctive decomposition. \cite{disj_dec_cautious_bold} proposed linear evidence based methods that only works in two particular cases, namely the \textit{consonant} and \textit{quasi-Bayesian} cases.
However, in the general case, the conjunctive and disjunctive decompositions cannot be simplified into mathematical expressions that would only feature focal sets. To the best of our knowledge, the only algorithms to have been proposed for this general computation are the FMT and matrix calculus \cite{matrix_cal}, as suggested in \cite{conj_dec,disj_dec_cautious_bold}, the latter being less efficient both in time and space than the former.
Thus, until now, all method based on these decompositions in the general case had to use the FMT, and so had a complexity at least exponential in time and space.

Here, we provide an evidence based method for the computation of the conjunctive and disjunctive decompositions in the general case. 
This paper is organized as follows : Section \ref{gretsi:preliminaries} introduces elements of DST on which we build our proposition. Section \ref{gretsi:methods} presents our method for the conjunctive decomposition and mathematically demonstrates why its complexity scales with information instead of $|\Omega|$. Section \ref{gretsi:methods_dual} transposes this method to the computation of the disjunctive decomposition. We then conclude in section \ref{gretsi:conclusion}.

\section{Preliminary definitions}\label{gretsi:preliminaries}

Before diving into the details of our method, it is necessary to recall some notions around the conjunctive decomposition of evidence. The disjunctive decomposition simply being its dual, what is said here for the conjunctive decomposition can easily be transposed to the former. We assume the reader is already familiar with the notions of frame of discernment (i.e. $\Omega$) and mass function.

\begin{definition}[\textit{Focal element}]
	For any mass function $m$, a focal element (a.k.a focal set) is a set $A$ such that $m(A) \neq 0$. 
	Sets that are not focal sets do not carry any information about $m$.
	In the following, we note $\mathcal{F}$ the set containing all focal sets of a mass function $m$.
\end{definition}

\subsection{Representations for a conjunctive fusion}

\begin{definition}[\textit{Commonality function}]
	For any mass function $m$, a commonality function $q$
	is defined as follows:
	\begin{align}\label{gretsi:q}
	\forall A \subseteq\Omega,\quad q(A) = \sum_{B\supseteq A}~ m(B) = \sum_{\substack{B\supseteq A\\B \in \mathcal{F}}}~ m(B)
	\end{align}
	
	It is important to notice that all subset of $\Omega$ may be associated with a commonality in $[0, 1]$, whether it is a focal set or not.
	
	\begin{remark}
		A procedure directly based on this formula would have a complexity between $O(|\mathcal{F}|)$ and $O(|\mathcal{F}|^2)$.
		Then, the mass function $m$ can simply be retrieved from $q$ with the same complexity by reversing the computation of (\ref{gretsi:q}) on focal sets only. However, this reverse computation assumes that focal sets are known, either because no modification has been done to $q$ or because of the use of well known operators like Dempster's fusion operator $\bigoplus$ that is known to only create focal sets at the intersection of the focal sets of the two previous mass functions. This also assumes that $m$ is always kept in memory or re-computed from $q$ to compute new values of $q$ on-the-fly. Otherwise, the reverse operation has to be performed by a general algorithm such as the FMT.
	\end{remark}
\end{definition}

\begin{definition}[\textit{Conjunctive fusion rule}]
	To fuse the information brought by two different sources $1$ et $2$, it is necessary to define a combination rule.
	The conjunctive fusion rule has been introduced by Smets 
	in the Transferable Belief Model (TBM) \cite{smets94}, through the binary operator $\rulecirc{$\cap$}$. It is designed to fuse two sources when both are considered reliable. The result is a conjunction of their statements.
	In commonality space, the conjunctive fusion rule is defined as follows:
	$$
	\forall A \subseteq \Omega,\quad (q_{1} \rulecirc{$\cap$} q_{2})(A) = q_1(A) ~.~ q_2(A)
	$$
\end{definition}

\begin{definition}[\textit{Conjunctive decomposition}]\label{gretsi:conj_weight}
	The conjunctive decomposition has been introduced in its general form by Smets \cite{conj_dec}.
	It decomposes any \textit{non-dogmatic} mass function, i.e. any mass function such that $\Omega \in \mathcal{F}$, into the conjunctive fusion of simple mass functions noted $A^w$, where $A \subset \Omega$ and $w$ is the conjunctive weight function defined thereafter. 
	
	In commonality space, we have:
	\begin{align*}
	\forall A \subset \Omega,\quad A_q^w : 
	\begin{cases}
	q(B) = 1			&\text{ $\forall B \subseteq A$ }\\
	q(B) = w(A)		&\text{ $\forall B \not\subseteq A$}
	\end{cases}
	\end{align*}
	The conjunctive decomposition of $q$ is:
	\begin{align*}
	q = \rulecirc{$\cap$}_{A\subset \Omega} A_q^{w}
	\end{align*}
	We get:
	\begin{align}\label{gretsi:conj_back_q}
	\forall B \subseteq \Omega,\quad q(B) 
	&= \prod_{\substack{A \not\supseteq B}} w(A) = q(\Omega).\prod_{\substack{A\subset \Omega\\A \supseteq B}} w(A)^{-1}
	\end{align}
	
	
	In consequence, according to the M\"obius inversion theorem recalled in \cite{FMT}, we have that for any mass function on $2^\Omega$, if $\Omega$ is a focal element (enables in particular to pose $w(\Omega)^{-1} = q(\Omega)$), then its conjunctive decomposition is defined by the following weight function $w$:
	\begin{align}\label{gretsi:conj_w}
	\forall B \subseteq\Omega,\quad w(B) = \prod_{\substack{A\supseteq B}}~ q(A)^{{(-1)}^{|A| - |B| + 1}}
	\end{align}
	
	We see that this weight function $w$ is not directly based on $m$ but on $q$, which does not possess any neutral value for non-focal elements. Moreover, replacing these commonalities by their equivalent sum of masses does not simply well in the general case. 
	
	\begin{remark}
		All in all, three points prevent an evidence based computation:
		\begin{itemize}
			\item the product on all supersets of $A$ in Eq. (\ref{gretsi:conj_back_q}), for all $B \in 2^\Omega$,
			\item same problem in Eq. (\ref{gretsi:conj_w}), 
			\item the ignorance about focal sets in $m$ if we modify $w$ or $q$ (e.g. through any combination rule other than Dempster's) to reverse Eq. (\ref{gretsi:q}).
		\end{itemize}
		If we could find a link between the focal elements of $m$ and the ones of $w-1$, we would be able to reduce the complexity of these transformations.
	\end{remark}
\end{definition}

\subsection{Representations for a disjunctive fusion (same remarks)}

\begin{definition}[\textit{Implicability function}]
	For any mass function $m$, an implicability function $b$
	is defined as follows:
	\begin{align}\label{gretsi:b}
	\forall A \subseteq\Omega,\quad b(A) = \sum_{B\subseteq A}~ m(B) = \sum_{\substack{B\subseteq A\\B \in \mathcal{F}}}~ m(B)
	\end{align}
	
	It is important to notice that all subset of $\Omega$ may be associated with an implicability in $[0, 1]$, whether it is a focal set or not.
\end{definition}

\begin{definition}[\textit{Disjunctive fusion rule}]
	The disjunctive fusion rule \cite{disj_rule} in an alternative fusion rule that is specific to DST. It exploits the existence of sets of different cardinalities in order to reflect a lack of trust in the sources to be fused. It corresponds to the disjunction of their statements, e.g. if one indicates an outcome $\{\omega_1\}$, while the other points to another outcome $\{\omega_2\}$, the resulting statement will be $\{\omega_1, \omega_2\}$ instead of the conjunctive result $\emptyset$.
	In implicability space, the disjunctive fusion rule is defined as follows:
	$$
	\forall A \subseteq \Omega,\quad (b_{1} \rulecirc{$\cup$} b_{2})(A) = b_1(A) ~.~ b_2(A)
	$$
\end{definition}

\begin{definition}[\textit{Disjunctive decomposition}]\label{gretsi:disj_weight}
	The disjunctive decomposition has been introduced in \cite{disj_dec_cautious_bold}.
	It decomposes any \textit{subnormal} mass function, i.e. any mass function such that $\emptyset \in \mathcal{F}$, into the disjunctive fusion of simple mass functions noted $A_v$, where $A \supset \emptyset$ and $v$ is the disjunctive weight function defined thereafter. 
	
	In implicability space, we have:
	\begin{align*}
	\forall A \supset \emptyset,\quad A^b_v : 
	\begin{cases}
	b(B) = 1			&\text{ $\forall B \supseteq A$ }\\
	b(B) = v(A)		&\text{ $\forall B \not\supseteq A$}
	\end{cases}
	\end{align*}
	The disjunctive decomposition of $b$ is:
	\begin{align*}
	b = \rulecirc{$\cup$}_{A\supset \emptyset} A^b_{v}
	\end{align*}
	We get:
	\begin{align}\label{gretsi:disj_back_b}
	\forall B \subseteq \Omega,\quad b(B) 
	&= \prod_{\substack{A \not\subseteq B}} v(A) = b(\emptyset).\prod_{\substack{A\supset \emptyset\\A \subseteq B}} v(A)^{-1}
	\end{align}
	
	
	In consequence, according to the M\"obius inversion theorem, we have that for any mass function on $2^\Omega$, if $\emptyset$ is a focal element (enables in particular to pose $v(\emptyset)^{-1} = b(\emptyset)$), then its disjunctive decomposition is defined by the following weight function $v$:
	\begin{align}\label{gretsi:disj_v}
	\forall B \subseteq\Omega,\quad v(B) = \prod_{\substack{A\subseteq B}}~ b(A)^{{(-1)}^{|A| - |B| + 1}}
	\end{align}
\end{definition}

\section{Evidence based computation of the conjunctive decomposition}\label{gretsi:methods}

In this section, we detail our evidence based method for the computation of the conjunctive decomposition. It exploits on the notion of \textit{focal point} that we introduce here, a notion derived from the one of focal set and that contains it. The dual of our method computing the disjunctive decomposition is deduced in Section \ref{gretsi:methods_dual} from the one for the conjunctive one, since the lattice $2^\Omega$ is symmetrical for any set $\Omega$.

More specifically, we provide a proof for what we call 
the \textit{proxy theorem} \ref{gretsi:GPT} and for its corollary \ref{gretsi:GNC}, which enable us to conclude that focal points are sufficient to the definition of the weight function $w$.
As a consequence, we propose a recursive formula using focal points that allows us to efficiently compute the conjunctive decomposition by reusing previously computed weights.




\subsection{Focal points}

\begin{definition}[\textit{Focal point}]\label{gretsi:FP_def}
	We define a \textit{focal point} as the set among all sets associated with a same commonality expression (i.e. a selection of focal supersets) that contains all the others. 
	Formally, noting $\mathring{\mathcal{F}}$ the set containing all focal points, they are defined by: 
	\begin{align*}
	\forall A \in \mathring{\mathcal{F}},~\forall B \subseteq \Omega,\quad \mathcal{F}_{\supseteq A} = \mathcal{F}_{\supseteq B} \Rightarrow A \supseteq B
	\end{align*}
	with the intuitive notation $\mathcal{F}_{\supseteq A} = \{F \in \mathcal{F} ~/~ F \supseteq A \}$.
	
	Thus, a focal point is the biggest set contained by all the sets in a selection of focal sets. In other words, it is the intersection of all these sets. 
	So, any focal point $A$ is defined by:
	$$\exists \mathcal{E} \subseteq \mathcal{F}, \mathcal{E} \neq \emptyset,\quad A = \bigcap_{F \in \mathcal{E}} F$$
\end{definition}

\begin{property}[\textit{Inclusion of focal sets}]\label{gretsi:inclusion}
	We have $\mathcal{F} \subseteq \mathring{\mathcal{F}}$ since 
	for all focal set $A \in \mathcal{F}$, there is a selection of focal sets $\mathcal{E} \subseteq \mathcal{F}$ such that $|\mathcal{E}| = 1$ and $A = \bigcap_{F \in \mathcal{E}} F$.
\end{property}

\begin{property}[\textit{Decomposition into bigger focal points}]\label{gretsi:sup_dec}
	The intersection of two focal points is also a focal point. It is equal to the union of their respective selections of focal sets:
	\begin{align*}
	\forall A, B \in \mathring{\mathcal{F}},\quad A \cap B = \left(\bigcap_{F \in \mathcal{F}_{\supseteq A}} F \right) \cap \left(\bigcap_{F \in \mathcal{F}_{\supseteq B}} F \right) = \bigcap_{F \in \mathcal{F}_{\supseteq A} \cup \mathcal{F}_{\supseteq B}} F
	\end{align*}
	\begin{remark}
		Note that this means any focal point can be described as either $F$, where $F \in \mathcal{F}$, or $\mathring{F} \cap F$, where $\mathring{F} \in \mathring{\mathcal{F}}$.
		Doing so, we can find all focal points of a belief assignment in $O\left(|\mathring{\mathcal{F}}|.|\mathcal{F}|\right)$.
	\end{remark}
	
\end{property}

\subsection{Computation of the conjunctive weight function $w$}

In this section, a first compact formula exploiting focal points to compute the weight function $w$ is provided in \autoref{gretsi:FPF}. Then, \autoref{gretsi:GPT} highlights an interesting property of this formula, which allows notably to prove \autoref{gretsi:GNC}, i.e. the sufficiency of focal points to define $w$. Finally, a recursive formula is given in \autoref{gretsi:RFPF}, which is more efficient than the one of \autoref{gretsi:FPF}.

\begin{lemma}[\textit{Exponent equilibrium}]\label{gretsi:exp_eq}
	Following the binomial theorem, we have $\forall C \subseteq \Omega$, $\forall A \subset C$:
	\begin{align*}
	\sum_{\substack{B\supseteq A\\B \subseteq C}}~ {(-1)}^{|B| - |A| + 1} = -\sum_{k=0}^{|C|-|A|}~ \binom{|C|-|A|}{k} {(-1)}^{k} .~ 1^{|C|-|A|-k} = -(1-1)^{|C|-|A|} = 0
	\end{align*}
\end{lemma}

\begin{theorem}[\textit{Focal point formula}]\label{gretsi:FPF}
	For any $A \subseteq \Omega$, the weight $w(A)$ can be decomposed into a product of commonalities on focal points supersets of $A$ only. Formally, we have:
	\begin{align}\label{gretsi:general_w}
	w(A) = q(A)^{-1}.\prod_{F \in \mathring{\mathcal{F}}_{\supset A}} q(F)^{e_{A, F}}
	\end{align} 
	with the intuitive notation $\mathring{\mathcal{F}}_{\supset A} = \{F \in \mathring{\mathcal{F}} ~/~ F \supset A \}$, and $$\forall F \in \mathring{\mathcal{F}}_{\supset A},\quad e_{A, F} =
	1-\displaystyle\sum_{\substack{B \in \mathring{\mathcal{F}}_{\supset A}\\B \subset F}} e_{A, B} 
	.$$
\end{theorem}
\begin{proof}
	First, let us note $e_{A, B}$ the sum of all exponents associated with a same commonality expression, defined by $\mathcal{F}_{\supseteq B}$, among sets containing $A$, excepted $A$ itself. More clearly, we pose $e_{A, B}$ as:
	$$e_{A, B} = \sum_{\substack{X\supset A\\\mathcal{F}_{\supseteq X} = \mathcal{F}_{\supseteq B}}}~ {(-1)}^{|X| - |A| + 1}$$
	
	Using Definition \ref{gretsi:FP_def}, we know that for any focal point $F \in \mathring{\mathcal{F}}_{\supset A}$, we have:
	$$e_{A, F} = \sum_{\substack{X\supset A\\X\subseteq F\\\mathcal{F}_{\supseteq X} = \mathcal{F}_{\supseteq F}}}~ {(-1)}^{|X| - |A| + 1}$$
	Moreover, note that for any couple of focal points $B, F \in \mathring{\mathcal{F}}$, if $B \neq F$, then $\mathcal{F}_{\supseteq B} \neq \mathcal{F}_{\supseteq F}$. So, for any set $A \subseteq \Omega$ and for any focal point $F \in \mathring{\mathcal{F}}$, we can introduce the following decomposition:
	\begin{align*}
	\prod_{\substack{X \supseteq A\\X \subseteq F}} q(X)^{(-1)^{|X|-|A|+1}} &= q(A)^{-1}.\left[\prod_{\substack{B \in \mathring{\mathcal{F}}_{\supset A}\\B \subset F}} q(B)^{e_{A,B}} \right].~q(F)^{e_{A,F}}
	\end{align*}
	This translates in terms of exponents as:
	\begin{align*}
	\sum_{\substack{X \supseteq A\\X \subseteq F}} (-1)^{|X|-|A|+1} &= -1 +\left[\sum_{\substack{B \in \mathring{\mathcal{F}}_{\supset A}\\B \subset F}} e_{A,B} \right] + e_{A,F},
	\end{align*}
	which is equal to 0, in virtue of Lemma \ref{gretsi:exp_eq}. This finally gives us:
	\begin{align*}
	e_{A,F} &= 1 -\sum_{\substack{B \in \mathring{\mathcal{F}}_{\supset A}\\B \subset F}} e_{A,B}
	\end{align*}
\end{proof}

\begin{example}[\textit{Quasi-Bayesian case}]\label{gretsi:quasi_bayes}
	Let us consider the case of a \textit{quasi-Bayesian} mass function $m$, i.e. a mass function such that $\Omega \in \mathcal{F}$ and all pairs of focal sets $(F_i, F_j)$, where $\Omega$ is neither of them, verify $F_i \cap F_j = \emptyset$. In this case, the only set that can be generated by the intersection of any number of focal points is $\emptyset$. Thus, we have $\mathring{\mathcal{F}} = \mathcal{F} \cup \{\emptyset\}$.
	Theorem \ref{gretsi:FPF} gives us for all set $A$ in $2^\Omega$:
	\begin{align}
	\forall A \subseteq \Omega,\quad w(A) = \begin{cases}
	q(\Omega)^{-1}		&\text{ if $A = \Omega$}\\
	q(A)^{-1} . ~q(\Omega)	&\text{ if $A \in \mathcal{F}\backslash\{\Omega\}$}\\
	q(A)^{-1} . ~q(\Omega)^{1-|\mathcal{F}\backslash\{\Omega\}|} . \displaystyle\prod_{\substack{F \in \mathcal{F}\backslash\{\Omega\}}} q(F) &\text{ if $A = \emptyset$}\\
	1	&\text{ otherwise}
	\end{cases},
	\end{align}
	where $q(\emptyset) = \sum_{B \in 2^\Omega} m(B) = 1$.
	This particular case was already known by \cite{disj_dec_cautious_bold} (Proposition 1).
\end{example}

\begin{theorem}[\textit{Proxy theorem}]\label{gretsi:GPT}
	For any set $A \subseteq \Omega$, if there is a smallest focal point superset, i.e. if there exists a focal point $P \in \mathring{\mathcal{F}}_{\supset A}$ such that for all focal points $F \in \mathring{\mathcal{F}}_{\supset A}$, we have $P \subseteq F$, then: 
	$$w(A) = q(A)^{-1}.~q(P).$$
\end{theorem}

\begin{proof}
	According to Theorem \ref{gretsi:FPF}, for any set $A \subseteq \Omega$, if there exists a focal point $P \in \mathring{\mathcal{F}}_{\supset A}$ such that for all focal point $F \in \mathring{\mathcal{F}}_{\supset A}$, we have $P \subseteq F$, then $e_{A,P} = 1$. Thus, for all focal point $F \in \mathring{\mathcal{F}}_{\supset P}$ such that there is no focal point in the open interval $(P, F)$, we have:
	$$e_{A,F} = 1 - e_{A,P} = 0.$$ 
	Similarly, for all focal point $F \in \mathring{\mathcal{F}}_{\supset P}$ such that, for all focal point $B$ in the open interval $(P, F)$, there is a no focal point in the open interval $(P, B)$,
	we have:
	$$e_{A,F} = 1 -(e_{A,P} + \sum_{\substack{B \in \mathring{\mathcal{F}}_{\supset P}\\B \subset F}} e_{A,B}) \\= 1 -(1 + \sum_{\substack{B \in \mathring{\mathcal{F}}_{\supset P}\\B \subset F}} 0) = 0.$$
	It is easy to see that by recursion, for all focal point $F \in \mathring{\mathcal{F}}_{\supset P}$, we obtain: $e_{A,F} = 0$.
	So, still according to Theorem \ref{gretsi:FPF}, we get:
	\begin{align*}
	w(A) = q(A)^{-1}.~ q(P)^1 ~. \prod_{F \in \mathring{\mathcal{F}}_{\supset P}} q(F)^0 = q(A)^{-1}. ~q(P)
	\end{align*}
\end{proof}

\begin{corollary}[\textit{Sufficiency of $\mathring{\mathcal{F}}$ to define $w$}]\label{gretsi:GNC}
	
	For all set $A$ that is not a focal point, there is a proxy focal point $P \in \mathring{\mathcal{F}}_{\supset A}$ such that $q(A) = q(P)$ and for all focal point $F \in \mathring{\mathcal{F}}_{\supset A}$, we have $P \subseteq F$. Thus, according to Theorem \ref{gretsi:GPT}, we have: $$\forall A \not\in \mathring{\mathcal{F}},\quad w(A) = 1.$$
\end{corollary}
\begin{proof} 
	By Definition \ref{gretsi:FP_def} of a focal point, for any set $A \subseteq \Omega$, there is a focal point $P \in \mathring{\mathcal{F}}_{\supseteq A}$ such that $\mathcal{F}_{\supseteq P} = \mathcal{F}_{\supseteq A} \Rightarrow P \supseteq A$. This means that $q(A) = q(P)$. Also, if $A$ is not a focal point, then $P \in \mathring{\mathcal{F}}_{\supset A}$ (since $\Omega \in \mathring{\mathcal{F}}$). In addition, if $A$ is not a focal point, Property \ref{gretsi:sup_dec} tells us that $$\bigcap \mathring{\mathcal{F}}_{\supset A} = \bigcap \left(\bigcup_{F\in \mathring{\mathcal{F}}_{\supset A}} \mathcal{F}_{\supseteq F}\right) = \bigcap \mathcal{F}_{\supset A} = P.$$ 
	By definition of the operator $\cap$, we get that $\forall F \in \mathring{\mathcal{F}}_{\supset A},~ F \supseteq P$.
\end{proof}

\begin{example}[\textit{Consonant case}]\label{gretsi:consonant}
	Let us now examine the case of a \textit{consonant} mass function, i.e. a mass function such that all its $n$ focal sets $F_i$ verify $F_1 \subset F_2 \subset \cdots \subset F_n \subset \Omega$. In this case, the proxy theorem \ref{gretsi:GPT} applies to all set $A$ in $2^\Omega\backslash\{\Omega\}$. Furthermore, as the intersection of any number of these focal sets is necessarily one of them, we have $\mathring{\mathcal{F}} = \mathcal{F}$. We obtain:
	\begin{align}
	\forall A \subseteq \Omega,\quad w(A) = \begin{cases}
	q(\Omega)^{-1}		&\text{ if $A = \Omega$}\\
	q(F_i)^{-1} . ~q(F_{i+1})	&\text{ if $\exists i \in \llbracket 1, n \rrbracket ~/~ A = F_i$}\\
	1	&\text{ otherwise}
	\end{cases}
	\end{align}
	This particular case was already known by \cite{disj_dec_cautious_bold} (Proposition 2).
\end{example}

\begin{theorem}[\textit{Recursive focal point formula}]\label{gretsi:RFPF}
	Thanks to Corollary \ref{gretsi:GNC} and taking back Eq. (\ref{gretsi:conj_back_q}), we obtain that for any set $A \subseteq \Omega$, the weight $w(A)$ can be decomposed into a product of weights associated with focal points supersets of $A$ as follows:
	\begin{align}\label{gretsi:w_general}
	\forall A \subseteq \Omega,\quad w(A) = \begin{cases}
	q(A)^{-1} . \prod_{\substack{F \in \mathring{\mathcal{F}}_{\supset A}}} w(F)^{-1}	&\text{ if $A \in \mathring{\mathcal{F}}$}\\
	1	&\text{ otherwise}
	\end{cases}
	\end{align}
	
	\begin{remark}
		A procedure directly based on this formula would have a complexity between $O(|\mathring{\mathcal{F}}|)$ and $O(|\mathring{\mathcal{F}}|^2)$.
		Its inverse (\ref{gretsi:conj_back_q}) only on focal points leads to the same complexity.
	\end{remark}
\end{theorem}





Thus, our method varies from $O(|\mathring{\mathcal{F}}|)$ to $O(|\mathring{\mathcal{F}}|^2)$, where $|\mathring{\mathcal{F}}| \in [~|\mathcal{F}|, ~2^{|\Omega|}~]$, depending on the structure of $\mathcal{F}$, as seen in the two notable particular cases presented in Examples \ref{gretsi:quasi_bayes} and \ref{gretsi:consonant}. In general, the more there are big focal sets that are not nested in others, the more $|\mathring{\mathcal{F}}|$ has chances to be great. Nevertheless, the most natural and interpretable structures are the ones close to either Example \ref{gretsi:quasi_bayes} or \ref{gretsi:consonant}, which tend to have a number of focal points close to their number of focal sets. Moreover, our approach directly benefits from methods that approximate $\mathcal{F}$, which should maintain $|\mathring{\mathcal{F}}|$ close to $|\mathcal{F}|$. These approximation methods do not limit the expressiveness of belief assignments.

\section{Transposition to the computation of the disjunctive decomposition}\label{gretsi:methods_dual}

In this section, with simply provide the dual of the theorems and corollaries of section \ref{gretsi:methods} that come from the symmetry of $2^\Omega$.

\subsection{Dual focal points}

\begin{definition}[\textit{Dual focal point}]\label{gretsi:FP_def_dual}
	We define a \textit{dual focal point} as the set among all sets associated with a same implicability expression (i.e. a selection of focal subsets) that is contained in all the others. 
	Formally, noting $\overline{\mathring{\mathcal{F}}}$ the set containing all dual focal points, they are defined by: 
	\begin{align*}
	\forall A \in \overline{\mathring{\mathcal{F}}},~\forall B \subseteq \Omega,\quad \mathcal{F}_{\subseteq A} = \mathcal{F}_{\subseteq B} \Rightarrow A \subseteq B
	\end{align*}
	with the intuitive notation $\mathcal{F}_{\subseteq A} = \{F \in \mathcal{F} ~/~ F \subseteq A \}$.
	
	Thus, a dual focal point is the smallest set containing all the sets in a selection of focal sets. In other words, it is the union of all these sets. 
	So, any dual focal point $A$ is defined by:
	$$\exists \mathcal{E} \subseteq \mathcal{F}, \mathcal{E} \neq \emptyset,\quad A = \bigcup_{F \in \mathcal{E}} F$$
\end{definition}

\begin{property}[\textit{Inclusion of focal sets}]\label{gretsi:inclusion_dual}
	As for focal points, we have $\mathcal{F} \subseteq \overline{\mathring{\mathcal{F}}}$ since 
	for all focal set $A \in \mathcal{F}$, there is a selection of focal sets $\mathcal{E} \subseteq \mathcal{F}$ such that $|\mathcal{E}| = 1$ and $A = \bigcup_{F \in \mathcal{E}} F$.
\end{property}

\begin{property}[\textit{Decomposition into smaller focal points}]\label{gretsi:inf_dec}
	The union of two dual focal points is also a dual focal point. It is equal to the union of their respective selections of focal sets:
	\begin{align*}
	\forall A, B \in \overline{\mathring{\mathcal{F}}},\quad A \cup B = \left(\bigcup_{F \in \mathcal{F}_{\supseteq A}} F \right) \cup \left(\bigcup_{F \in \mathcal{F}_{\supseteq B}} F \right) = \bigcup_{F \in \mathcal{F}_{\supseteq A} \cup \mathcal{F}_{\supseteq B}} F
	\end{align*}
	\begin{remark}
		Note that this means any dual focal point can be described as either $F$, where $F \in \mathcal{F}$, or $\mathring{F} \cup F$, where $\mathring{F} \in \overline{\mathring{\mathcal{F}}}$.
		Doing so, we can find all dual focal points of a belief assignment in $O\left(|\overline{\mathring{\mathcal{F}}}|.|\mathcal{F}|\right)$.
	\end{remark}
	
\end{property}

\subsection{Computation of the disjunctive weight function $v$}

\begin{corollary}[\textit{Dual focal point formula}]\label{gretsi:FPF_dual}
	For any $A \subseteq \Omega$, the weight $v(A)$ can be decomposed into a product of implicabilities on focal points subsets of $A$ only. Formally, we have:
	\begin{align}\label{gretsi:general_v}
	v(A) = b(A)^{-1}.\prod_{F \in \overline{\mathring{\mathcal{F}}}_{\subset A}} b(F)^{e_{A, F}}
	\end{align} 
	with the intuitive notation $\overline{\mathring{\mathcal{F}}}_{\subset A} = \{F \in \overline{\mathring{\mathcal{F}}} ~/~ F \subset A \}$, and $$\forall F \in \overline{\mathring{\mathcal{F}}}_{\subset A},\quad e_{A, F} =
	1-\displaystyle\sum_{\substack{B \in \overline{\mathring{\mathcal{F}}}_{\subset A}\\B \supset F}} e_{A, B} 
	.$$
\end{corollary}

\begin{example}[\textit{Quasi-Bayesian dual case}]\label{gretsi:quasi_bayes_dual}
	Let us consider the dual of the \textit{quasi-Bayesian} case of Example \ref{gretsi:quasi_bayes}. This means that we have a mass function $m$ such that $\emptyset \in \mathcal{F}$ and all pairs of focal sets $(F_i, F_j)$, where $\emptyset$ is neither of them, verify $F_i \cup F_j = \Omega$. In this case, the only set that can be generated by the union of any number of dual focal points is $\Omega$. Thus, we have $\overline{\mathring{\mathcal{F}}} = \mathcal{F} \cup \{\Omega\}$.
	Theorem \ref{gretsi:FPF_dual} gives us for all set $A$ in $2^\Omega$:
	\begin{align}
	\forall A \subseteq \Omega,\quad v(A) = \begin{cases}
	b(\emptyset)^{-1}		&\text{ if $A = \emptyset$}\\
	b(A)^{-1} . ~b(\emptyset)	&\text{ if $A \in \mathcal{F}\backslash\{\emptyset\}$}\\
	b(A)^{-1} . ~b(\emptyset)^{1-|\mathcal{F}\backslash\{\emptyset\}|} . \displaystyle\prod_{\substack{F \in \mathcal{F}\backslash\{\emptyset\}}} b(F) &\text{ if $A = \Omega$}\\
	1	&\text{ otherwise}
	\end{cases},
	\end{align}
	where $b(\Omega) = \sum_{B \in 2^\Omega} m(B) = 1$. Notice however that the quasi-Bayesian case of Example \ref{gretsi:quasi_bayes} is actually the worst case for the complexity tied to the computation of $v$. More precisely, the worst case is the one corresponding to $\mathcal{F}$ containing all singletons of $2^\Omega$. In the same way, this dual quasi-Bayesian case is the worst case complexity for the computation of $w$. More precisely, the worst for $w$ is attained when the complement to $\Omega$ of each singleton is a focal set.  
\end{example}

\begin{theorem}[\textit{Proxy theorem}]\label{gretsi:GPT_dual}
	For any set $A \subseteq \Omega$, if there is a biggest focal point subset, i.e. if there exists a dual focal point $P \in \overline{\mathring{\mathcal{F}}}_{\subset A}$ such that for all dual focal points $F \in \overline{\mathring{\mathcal{F}}}_{\subset A}$, we have $P \supseteq F$, then: 
	$$v(A) = b(A)^{-1}.~b(P).$$
\end{theorem}

\begin{corollary}[\textit{Sufficiency of $\overline{\mathring{\mathcal{F}}}$ to define $v$}]\label{gretsi:GNC_dual}
	
	For all set $A$ that is not a dual focal point, there is a proxy dual focal point $P \in \overline{\mathring{\mathcal{F}}}_{\subset A}$ such that $b(A) = b(P)$ and for all dual focal point $F \in \overline{\mathring{\mathcal{F}}}_{\subset A}$, we have $P \supseteq F$. Thus, according to Theorem \ref{gretsi:GPT_dual}, we have: $$\forall A \not\in \overline{\mathring{\mathcal{F}}},\quad v(A) = 1.$$
\end{corollary}

\begin{corollary}[\textit{Recursive dual focal point formula}]\label{gretsi:RFPF_dual}
	Thanks to Corollary \ref{gretsi:GNC_dual} and taking back Eq. (\ref{gretsi:disj_back_b}), we obtain that for any set $A \subseteq \Omega$, the weight $v(A)$ can be decomposed into a product of weights associated with dual focal points subsets of $A$ as follows:
	\begin{align}\label{gretsi:v_general}
	\forall A \subseteq \Omega,\quad v(A) = \begin{cases}
	b(A)^{-1} . \prod_{\substack{F \in \overline{\mathring{\mathcal{F}}}_{\subset A}}} v(F)^{-1}	&\text{ if $A \in \overline{\mathring{\mathcal{F}}}$}\\
	1	&\text{ otherwise}
	\end{cases}
	\end{align}
	
	\begin{remark}
		A procedure directly based on this formula would have a complexity between $O(|\overline{\mathring{\mathcal{F}}}|)$ and $O(|\overline{\mathring{\mathcal{F}}}|^2)$.
		Its inverse (\ref{gretsi:conj_back_q}) only on focal points leads to the same complexity.
	\end{remark}
\end{corollary}

\section{Conclusion and perspectives}\label{gretsi:conclusion}

We have presented here the difficulties tied to the design of evidence based algorithms for the transformation of a belief assignment into its conjunctive or disjunctive decomposition and its inverse. For this, we proposed a novel mathematical notion that we called \textit{focal point}, derived from the notion of focal set. With these focal points, we exploit the properties of the function to be transformed in $2^\Omega$, while the state-of-the-art optimal general algorithms, based on the \textit{Fast Möbius Transform} (FMT), ignore it. Doing so, we can now design new general algorithms with complexities inferior to the exponential one of these optimal algorithms. 
Outside the scope of this article, the notion of focal point can be used in DST for information analysis without having to compute the conjunctive (or disjunctive) decomposition beforehand. More generally, any method exploiting the conjunctive or disjunctive decomposition at some point, like the Cautious fusion rule or the Bold fusion rule, can benefit from a reduction in their complexity.

However, the computation of these focal points has a complexity in $O(|\mathring{\mathcal{F}}|~.~|\mathcal{F}|)$, where $\mathring{\mathcal{F}}$ is the set containing these focal points and $\mathcal{F}$ is the set containing the focal sets of some belief assignment. Depending on $\mathcal{F}$, the set $\mathring{\mathcal{F}}$ might be as big as $2^\Omega$, which means that $O(|\mathring{\mathcal{F}}|~.~|\mathcal{F}|)$ might be far greater than $O(|\Omega|~.~2^{|\Omega|})$, i.e. the complexity of the FMT, in some cases. Thus, we later provided complementary methods in \cite{me} guaranteeing better complexities than the FMT. Furthermore, the question of how to compute back the mass function from either the commonality or implicability function in the general case without the FMT remains. Ideally, we would like to exploit these already computed focal points once again, after modification of the conjunctive or disjunctive decomposition. This requires a broader view on our approach, which we later provided in \cite{me_journal}.

%

\bibliography{./complexity-reduction}

\end{document}